\documentclass{article}
\usepackage{graphicx} 

\title{On Functional Activations in Deep Neural Networks}
\author{Andrew S. Nencka, L. Tugan Muftuler, Peter LaViolette, Kevin M. Koch}
\date{\today}

\begin{document}

\maketitle

\section{Abstract}
\noindent \textbf{Background:} Deep neural networks have proven to be powerful computational tools for modeling, prediction, and generation. However, the workings of these models have generally been opaque. Recent work has shown that the performance of some models are modulated by overlapping functional networks of connections within the models. Here the techniques of functional neuroimaging are applied to an exemplary large language model to probe its functional structure.

\noindent \textbf{Methods:} A series of block-designed task-based prompt sequences were generated to probe the Facebook Galactica-125M model. Tasks included prompts relating to political science, medical imaging, paleontology, archeology, pathology, and random strings presented in an off/on/off pattern with prompts about other random topics. For the generation of each output token, all layer output values were saved to create an effective time series. General linear models were fit to the data to identify layer output values which were active with the tasks.

\noindent \textbf{Results:} Distinct, overlapping networks were identified with each task. Most overlap was observed between medical imaging and pathology networks. These networks were repeatable across repeated performance of related tasks, and correspondence of identified functional networks and activation in tasks not used to define the functional networks was shown to accurately identify the presented task.

\noindent \textbf{Conclusion:} The techniques of functional neuroimaging can be applied to deep neural networks as a means to probe their workings. Identified functional networks hold the potential for use in model alignment, modulation of model output, and identifying weights to target in fine-tuning.

\section{Introduction}

In recent years, the power of deep neural networks, particularly transformer models, has been  exemplified across a spectrum of domains \cite{vaswani2017attention, alzubaidi2021review}. Among these demonstrations, Large Language Models (LLMs) have grabbed the attention of researchers across disciplines due to the transformative potential of such architectures \cite{brown2020language}. However, as the complexity of these models scales, so does the non-transparent nature of their operational mechanics.

The opacity of large models coupled with their increasing power and functional relevance begets a pressing need for methods that elucidate their inner workings \cite{belinkov2019analysis}. The quest for understanding the intricacies of transformers is not merely an academic endeavor, but a pragmatic requisite to identify and ameliorate states leading to sub-optimal or undesirable performance.

The opaque box nature of large transformer models hinders not only the interpretability but also the controllability of the models \cite{wang2023aligning}. While it may be inconvenient to be unable to explain why a model succeeded in a given task, recent discussion of the dangers of model misalignment highlight the danger of being unable to clearly elucidate why a model failed in its task. The magnitude of this challenge continues to be amplified with the introduction of larger scaled models \cite{zhao2023survey}.   While such models are capable of astonishing feats \cite{xi2023rise}, their increasing size further obfuscates pathways to understanding the relationship between their structure and function.

Historically, "feature visualization" has served as a beacon in the murkiness of neural network interpretation \cite{olah2017feature}. By isolating and visualizing the impact of individual neurons, researchers have gleaned insights into the model's behavior. Feature visualization provided a quantum step forward in understanding the nature of deep neural networks. However, this method, while illuminative, is akin to perceiving trees in lieu of the forest when applied to the colossal and complex transformer models.

A noteworthy stride towards demystifying transformers has been recently publicized by researchers at Anthropic \cite{bricken2023monosemanticity} and pioneered by many \cite{belinkov2019analysis, olah2020zoom, elhage2022toy}, who have embarked on training neural networks to ferret out function-dedicated sub-networks of model parameters nestled within transformers. This transcends the conventional neuron-centric approach, instead considering the collective functionality of neuron cohorts. Such a network-scale approach offers a deeper insight into what the model is essentially doing. The notion of 'functional networks' burgeons the scope of interpretation by spotlighting the synergy among neurons as opposed to the effect of solitary neurons.

The consideration of networks of activations within a deep neural network presents a parallel to the field of neuroscience where there has been tremendous insight arising from cross-disciplinary analogues \cite{POLDRACK2008223}. Here, we present a method to assess the activation of sub-networks of a deep neural network as a response to a given task. This parallels the earliest work of functional neuroimaging where research participants were dynamically imaged in a magnetic resonance imaging system while being presented with a task in an off/on/off block design \cite{doi:10.1126/science.1948051, doi:10.1073/pnas.89.12.5675, bandettini1992time}.

The past three decades of development in the field of functional neuroimaging have yielded a significant number of finely tuned tasks for probing the opaque box of natural neural networks (mediated by confounding physiologic processes \cite{d'esposito2003alterations}) in exquisite detail \cite{rakhimberdina2021natural}. In this work we illustrate that the task-based block-designed functional neuroimgaing paradigm can be adapted to probe the inner-workings of a deep neural-network. In this way, the experimental design and interpretive frameworks that are central to the functional neuroimaging community can be extended to the task of interpreting the functionality of deep neural networks, making the opaque box of the network somewhat more transparent.

With a more interpretable understanding of the underlying mechanisms of transformers, numerous opportunities arise. With the functions of different sub-networks identified, it may be possible to steer the model from undesirable behaviors towards more desirable outcomes. It may also be possible to assess model performance based upon the activation of a set of sub-networks prior to the completion of an inference task, thereby offering a mechanism to modulate model output.

\section{Methods}

Herein we present a proof of concept of this task-based functional mapping of the elements embedded in a deep neural network. We provide a brief description of how a functional neuroimaging experiment is performed, and then describe how the functional neuroimaging experiment paradigm can be adapted to the assessment of computed layer outputs throughout the deep neural network. This paradigm is then applied to a proof of concept application of assessing sub-network activation in the Facebook Galactica-125M model \cite{taylor2022galactica} across a set of tasks spanning multiple academic fields.

\subsection{Functional Neuroimaging}

Shortly after the blood oxygenation level dependent (BOLD) contrast \cite{doi:10.1073/pnas.87.24.9868} was described, numerous labs used it with magnetic resonance imaging (MRI) to map regions of the brain involved with different functional tasks \cite{doi:10.1126/science.1948051, doi:10.1073/pnas.89.12.5675, bandettini1992time}. 

Task-based functional MRI (fMRI) experiments include presenting a research participant with a known stimulus which is often organized into a series of "on," and "off" blocks where the stimulus of interest is, or is not present. A time series of volumetric images of the brain are acquired while the task is presented, and each volume pixel (or voxel) of the brain is considered as a unique time series. A statistical model is fit to each imaged voxel time series to identify if the voxel's BOLD contrast changes are related to the presented task. If they are found to correlate with the task, the voxel is identified to be "active," and maps of active voxels across the brain are generated to identify the cortical network or networks associated with the presented task.

There are confounds in fMRI, however. These confounds include most obviously the spatial scale of the observed voxels and the underlying active neurons. Voxels with spatial scales on the orders of millimeters sample tissue covering thousands of neurons and a variety of tissues including neurons, glial cells, blood vessels, cerebrospinal fluid, and more \cite{dukart2014structure}. Less obviously, but more impactfully, the observed BOLD contrast is related to underlying neuronal activity through a cascading series of physiologic functions which, at best, modulates the observed signal to be a convolution of a physiologic response function to presented stimulus \cite{LINDQUIST2009S187}, or, at worst,  may become entirely decoupled from actual neuronal activity \cite{d'esposito2003alterations}.

\subsection{ Neural Network Activation}

Here it is posited that a deep neural network can be considered in a similar way to the brain in an fMRI experiment. As input data passes through the connections of a deep neural network, the values output from each node are input into following nodes as a product of the output value and the weight of the connection which is defined in training. One can consider each of these connections to be analogous to a voxel in an fMRI experiment.

As with a task-based fMRI experiment, the neural network can be probed with a series of inputs related to, or not related to, a specific task in an a priori known pattern. The values passed through the neural network connections can be saved for each input, and a time series of layer outputs can be generated for each inference step of the model. In the case of a transformer used for a LLM, each time point in the generated series can correspond to an output token. With the time series of layer outputs retained, the statistical modeling developed in fMRI can be applied to the layer output value time series to determine which layer output connections are preferentially activated with the presented task.

Interestingly, this analogy includes fewer confounds compared to the predicate neuroscience example. With each layer output value considered individually, this technique is not compromised by the spatial averaging present in brain imaging (though it also does not benefit from the dimensionality reduction afforded by such averaging). Additionally, with the task input explicitly interacting with the observed layer outputs, observed waveforms are not impacted by a physiologic response function (though attention mechanisms in a transformer may introduce a temporal autocorrelation).

\subsection{Implementation}

To simplify scaling, the Facebook Glactica-125M model (Meta, \cite{GALACTICA}) was considered due to its limited number of layer output values. The model was imported into a Jupyter notebook in an NVIDIA-Docker container (nvcr.io/nvidia/pytorch:23.08-py3) as a HuggingFace transformer \cite{wolf2020transformers}. Pytorch hooks to save output tensor values were added to each module in the model. For inference, a wrapper function was developed to save the computed layer output tensor on an output token-by-token basis.

A series of block designed experiments were designed to probe different potential sub-networks based upon varying knowledge fields: Political science (Pol. Sci.), Medical Imaging (Med. Img.), Paleontology (Paleo.), Archeology (Arch.), and Pathology (Path.). To probe these fields, Chat-GPT 4.0 \cite{openai_chat2023} was instructed to develop 100 prompts for the Glactica model for each of these fields, as well as six sets of 100 prompts which explicitly do not include the seven listed fields. These prompts were:
\begin{itemize}
	\item Please generate a set of 100 short prompts for the Facebook Galactica model on the topic of political science. Do not use any words more than 10 times in this list of prompts. Cover topics including forms of government, theory of government, citizen engagement, and other broad aspects of political science . Please return the results as a python list of strings. Again, DO NOT repeat words across prompts more than 10 times.
	\item Please generate a set of 100 short prompts for the Facebook Galactica model on the topic of medical imaging and radiology. Do not use any words more than 10 times in this list of prompts. Cover topics including the physics,  medical applications, modalities, contrast, acquisition, and reconstruction of medical images in radiology. Please return the results as a python list of strings. Again, DO NOT repeat words across prompts more than 10 times.
	\item Please generate a set of 100 short prompts for the Facebook Glactica model on the topic of paleontology. Do not use any words more than 10 times in this list of prompts. Cover topics including fosilization processes, types of fossils, evolutionary biology, methods and techniques of paleontology, mass extinctions, and the ancient environment. Please return the result as a python list of strings. Again, DO NOT repeat words across prompts more than 10 times.
	\item Please generate a set of 100 short prompts for the Facebook Glactica model on the topic of archeology. Do not use any words more than 10 times in this list of prompts. Cover topics including historical and prehistorical periods, field survey, excavation, site mapping, artifact analysis, archeometry, dating methods, lithic and ceramic analysis. Please return the result as a python list of strings. Again, DO NOT repeat words across prompts more than 10 times.
	\item Please generate a set of 100 short prompts for the Facebook Glactica model on the topic of pathology. Do not use any words more than 10 times in this list of prompts. Cover topics including anatomic and clinical pathology, biopsy, surgical sampling, sample staining, immunohistochemistry, forensics, cancer and disease, and diagnostics. Please return the result as a python list of strings. Again, DO NOT repeat words across prompts more than 10 times.
	\item Please generate six sets of 100 short prompts for the Facebook Glactica model spread across any topics OTHER THAN political science, medical imaging, paleontology, archeology, or pathology. Do not use any words more than 10 times in each list of prompts. Each list should include a wide assortment of topics. DO NOT INCLUDE topics of political science, medical imaging, paleontology, archeology, or pathology. Please return the result as five separate python lists of strings. Again, DO NOT repeat words across prompts in a list more than 10 times. The six lists should each contain a random selection of topics, and the lists should be titled `random\_prompts\_1', `random\_prompts\_2', `random\_prompts\_3', `random\_prompts\_4',  `random\_prompts\_5', and `random\_prompts\_6'.
\end{itemize}

The natural language processing toolkit (NLTK, \cite{nltk, nltkbook}) was used to generate an additional 100 strings of random words as an additional null control group.

\begin{figure}
    \centering
    \includegraphics[width=0.5\linewidth]{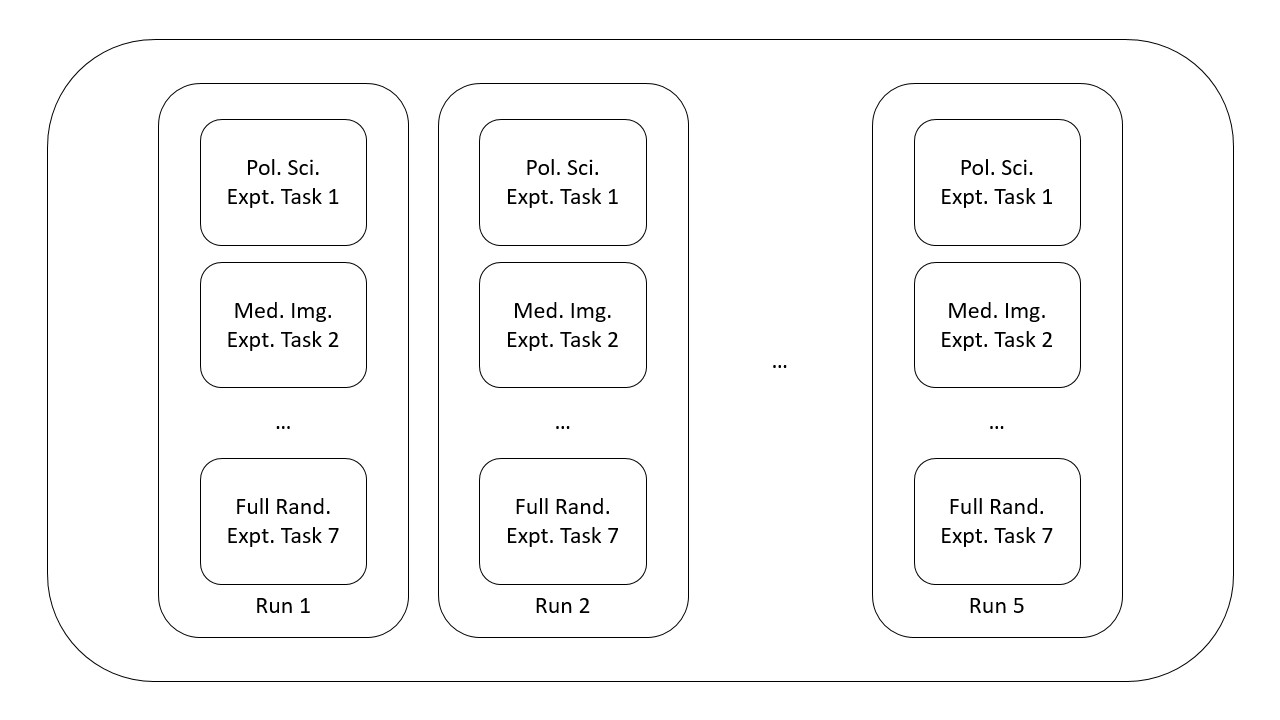}
    \caption{The design of the presented work includes five runs, with each including seven presented experimental tasks. Prompts for each task were different across runs, and were generated to encompass semantic content aligned with the named experimental tasks. Analysis included comparison of results between experiments within a single run, comparison of results for a given experiment across runs, and comparison of results for a held out experiment with the collective results across the remaining runs.}
    \label{exptDesign}
\end{figure}

Seven block designed experiments were created. Each experiment included a different task including: Political science versus Chat-GPT random prompts set 1, Medical imaging versus Chat-GPT random prompts set 2, Paleontology versus Chat-GPT random prompts set 3, Archeology versus Chat-GPT random prompts set 4, Pathology versus Chat-GPT random prompts set 5, NLTK random prompts versus Chat-GPT random prompts set 6, and Chat-GPT random prompts set 1 versus Chat-GPT random prompts set 2. Each block of each experiment included the input of one prompt from the specified set of prompts used to generate up to 10 tokens. Each experiment included eleven blocks of the "off" Chat-GPT generated random prompts and ten interleaved blocks of the listed task "on" blocks so that the experiment started and finished with an "off" block. This set of seven experiments was repeated in five runs. The organization of this design is shown in Fig. \ref{exptDesign}.

To simplify the process of saving layer outputs, inference was performed on the central processing unit. Inference was performed on a server with an Intel Xeon processor with 12 cores and 192 Gb of RAM. 

For each prompt, the above described wrapper function was used to call the LLM with a limit of 10 new tokens created for each model call. For each output token, layer outputs which were passed through the neural network to generate it were saved. The saved layer outputs were concatenated into an experiment-specific time series across all tokens generated in each experiment.

A simple linear model consisting of a baseline and a binary regressor (0 for "off" and 1 for "on" blocks) was fit to each layer output value time series. The model was fit using the statsmodels \cite{seabold2010statsmodels} general linear model (GLM, \cite{68aee965-a8a0-3e72-9f89-8d89ae91a62b}) function. With 259,744 layer output values saved with each generated token, a layer output was defined to be active with a Bonferroni corrected p=0.0001 threshold \cite{dunn1961multiple}. This model was fit independently for each layer output value across all experiments.


\section{Results}

Analysis was performed to assess if the LLM exhibits consistent task-specific functional networks. 

\begin{figure}
    \centering
    \includegraphics[width=0.5\linewidth]{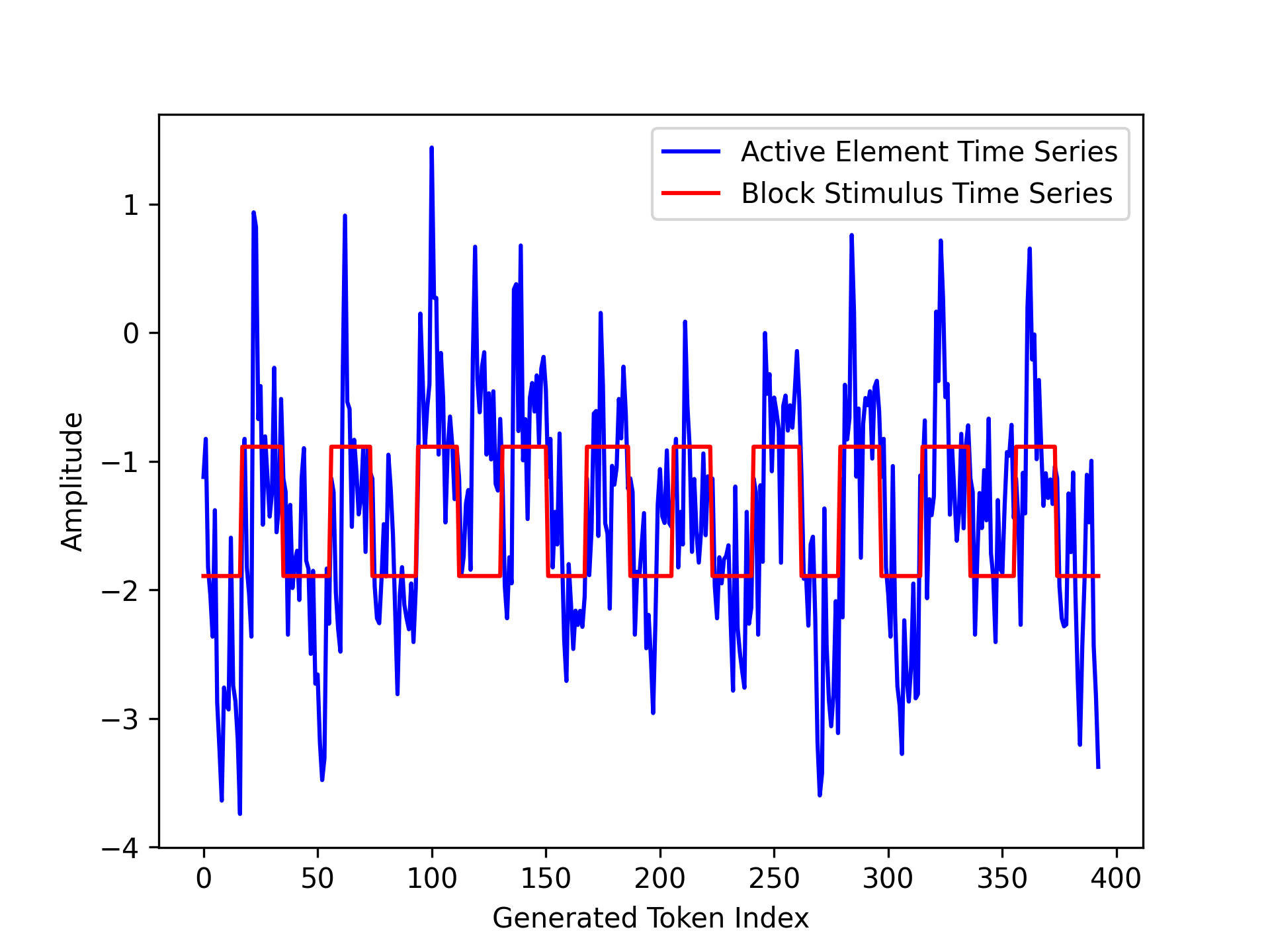}
    \caption{Example layer output value time series for an element that was identified to be active, blue, and the fit activation time series profile, red.}
    \label{repTS}
\end{figure}

An example time series from an active layer output is shown in Fig. \ref{repTS}. The observed active layer output is plotted in blue, while the fit block design waveform with eleven blocks of "off" and ten blocks of "on" is shown in red. The active and inactive blocks are readily apparent to the most casual observer, though there is clear variability of the layer output signal across the time series of inferred tokens.

\begin{figure}
    \centering
    \includegraphics[width=\linewidth]{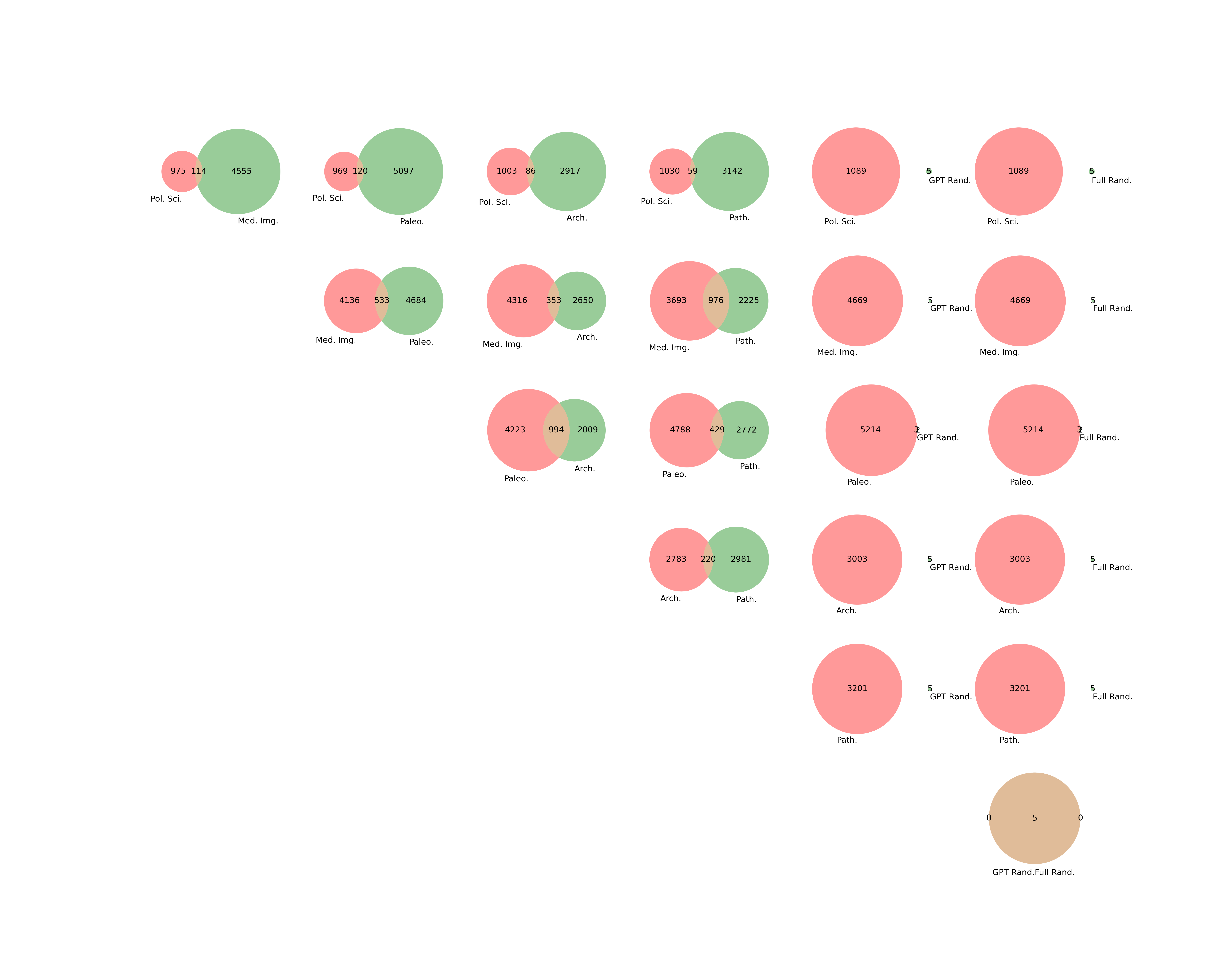}
    \caption{Venn diagrams showing the overlap of network activations across experimental tasks for an example run across the several task experiments. Experimental tasks include responding to prompts related to political science (Pol. Sci.), medical imaging (Med. Img.), paleontology (Paleo.), archeology (Arch.), pathology (Path.), random coherent prompts (GPT Rand.), or prompts of random words (Full Rand). There is some, but not much, overlap between active networks with different experimental tasks within a run. More overlap is observed between similar tasks (i.e. medical imaging and pathology, or paleontology and archeology) than dissimilar tasks (i.e. political science and paleontology).}
    \label{exptVenn}
\end{figure}

To assess the overlap of different functional networks, a series of Venn diagrams were considered across experiments in one run. The indices of layer output values which were found to be active in each task was compared with the indices of active layer output values in all other tasks in that run. These Venn diagrams are shown in Fig. \ref{exptVenn}. Labeling of the experimental tasks are abbreviated as described above, and the number of active layer outputs in each set are indicated on the plots. Unsurprisingly, activations with random inputs are quite limited. Interestingly, different tasks include significantly different numbers of activations--suggesting that some functional networks have representations which span greater numbers of observed layer output values. Suggestive of overlapping semantic representations, the greatest overlap was identified across the medical fields of medical imaging and pathology.

\begin{figure}
    \centering
    \includegraphics[width=\linewidth]{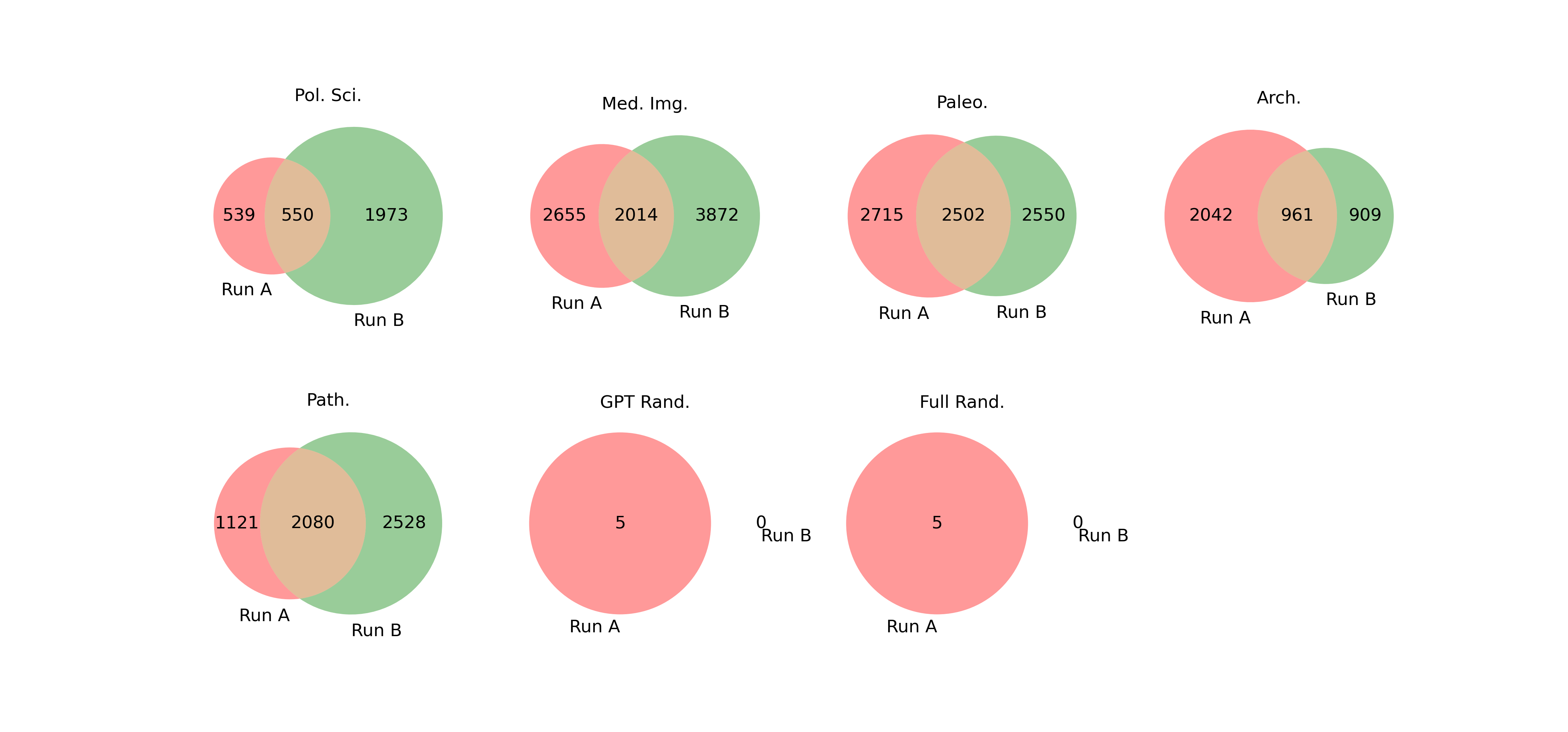}
    \caption{Venn diagrams illustrating the overlap of activations across runs for matched experimental tasks. Experimental tasks include responding to prompts related to political science (Pol. Sci.), medical imaging (Med. Img.), paleontology (Paleo.), archeology (Arch.), pathology (Path.), random coherent prompts (GPT Rand.), or prompts of random words (Full Rand). Significant overlap of active layer outputs are found for differing runs for each experimental task. This overlap is much greater than the overlap seen across tasks within a single run. }
    \label{runVenn}
\end{figure}

To assess the repeatability of the identified functional networks within the LLM, a series of Venn diagrams were considered across runs for each experiment. The indices of layer output values which were found to be active for a given experimental task in one run were compared with the indices of active layer outputs in another run for that experiment. These Venn diagrams are shown in Fig. \ref{runVenn}. Overlap within experimental task across runs is much greater than the overlap identified across tasks. The lack of total overlap is unsurprising as it is indicative of natural variation in network activation introduced by variance in provided prompts.

\begin{figure}
    \centering
    \includegraphics[width=\linewidth]{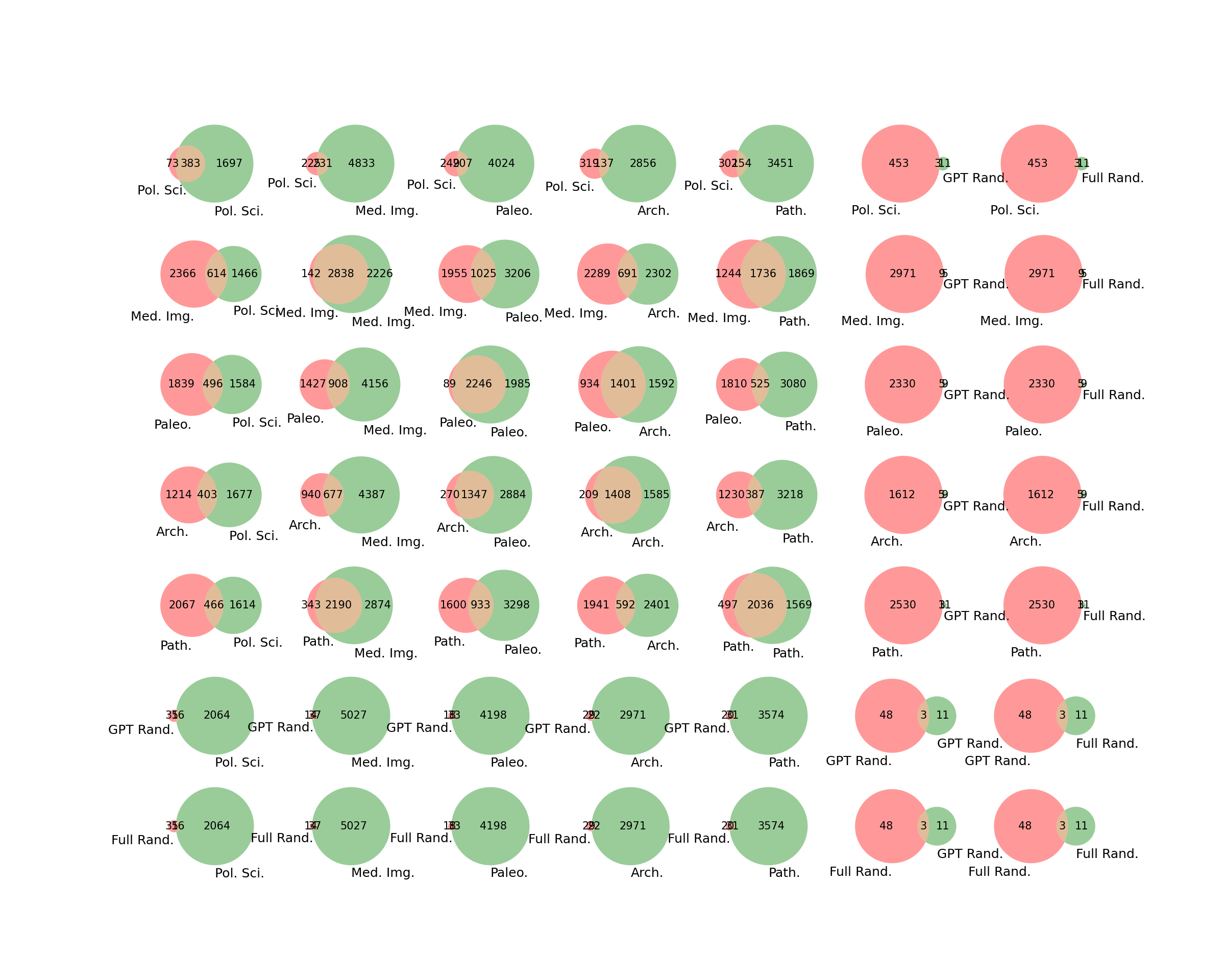}
    \caption{Venn diagrams showing the overlap of network activations for tasks in a held out run with the networks identified for each task across the other experimental runs. Each row corresponds to a presented task, and each column corresponds with a task-based functional network. Highest overlap of task activations with identified networks occurs when tasks align with the defined networks, as seen along the diagonal. High levels of overlap also exist between similar fields (i.e. medical imaging and pathology, or archaeology and paleontology. Tasks include requests for information regarding political science (Pol. Sci.), medical imaging (Med. Img.), paleontology (Paleo.), archaeology (Arch.), pathology (Path.), random prompts from Chat-GPT (GPT Rand.), and strings of random words (Full Rand.).}
    \label{exptVenn2}
\end{figure}

To assess the predictive power of the identified functional networks for political science, medical imaging, paleontology, archeology, and pathology, a network template based analysis was considered. One run of all seven experiments was segregated from the dataset. The set of layer outputs which were found to be active in at least three of the four remaining runs were identified for each experiment. These seven sets of layer outputs were defined to be the template functional networks associated with the seven tasks. Activations were computed on the experiments of the held out run. A set of Venn diagrams showing the overlap of the held out run with the defined functional networks is shown in Fig. \ref{exptVenn2}. 

\begin{figure}
    \centering
    \includegraphics[width=\linewidth]{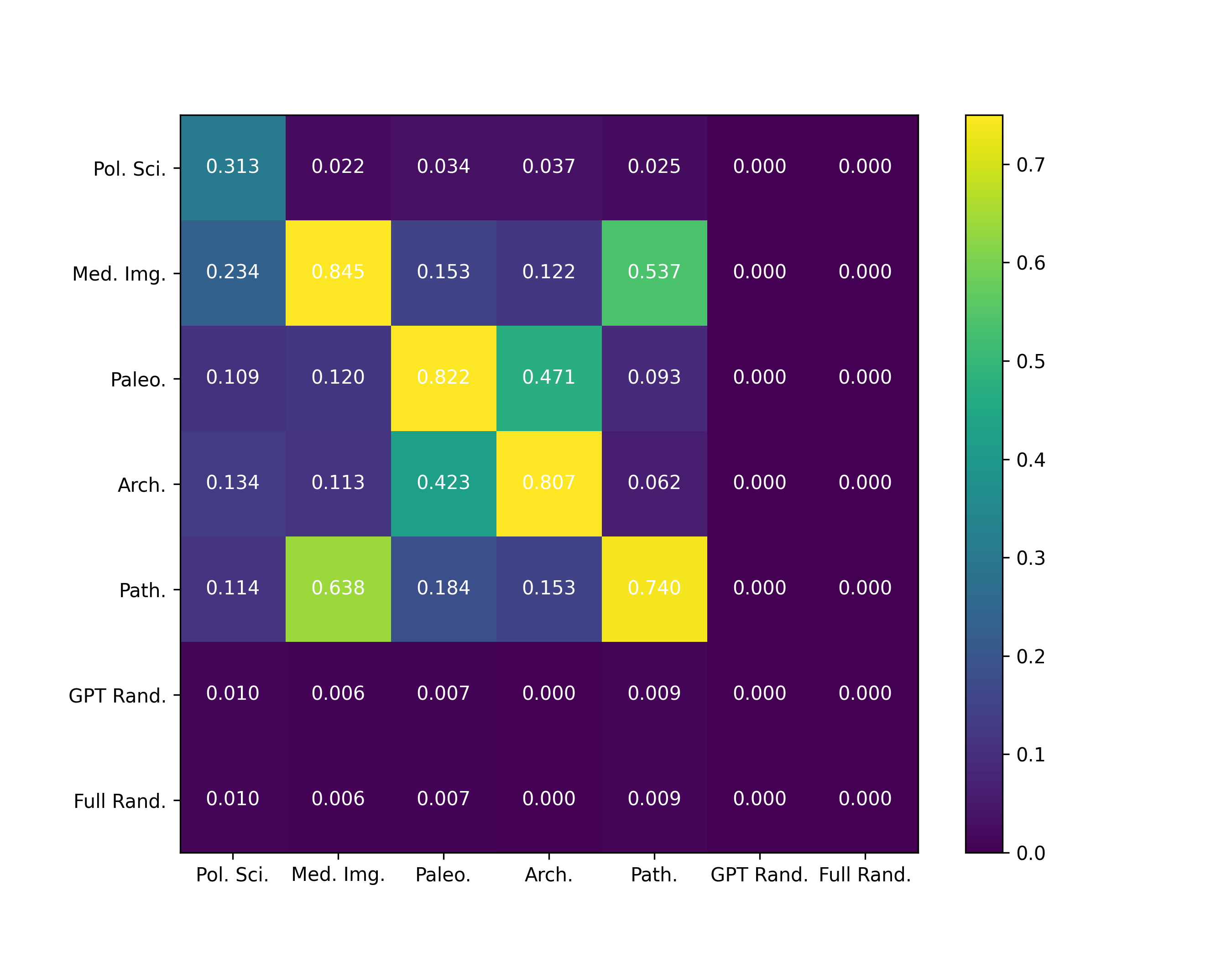}
    \caption{Percentage of functional network active for each held out experiment and each functional network. The high values on the diagonal indicate that this metric successfully identifies task based upon measured activations.}
    \label{predictiveAct}
\end{figure}

This overlap can be used to identify the presented experimental task in the held out run. The intersection of the set of run-specific activations with each functional network was computed and normalized by number of active layer outputs in the functional network. This percentage of functional network active metric is shown for each experimental run (row) and compared functional network (column) in Fig. \ref{predictiveAct}.

Unsurprisingly, random stimuli failed to yield any layer outputs which were consistently active across the template generation stage. This yielded functional networks of zero elements and undefined percentage of functional network active values in the right most columns and consistent values across both of the random input experiments (bottom rows). In cases where the functional networks were not empty, there is a clear correspondence of this metric being elevated when the experimental task aligns with the functional network template for that task. This is visualized by the elevated values along the diagonal of Fig. \ref{predictiveAct}. It is clear that, in this exemplary data, an arbitrary threshold where 70\% of the functional network being active would yield correct task identification in the cases of medical imaging, paleontology, archeology, and pathology tasks. Further, if the task identification process was selected to be the the network wherein this metric is the greatest across all networks, it would be successful in all except the random experimental tasks.

\section{Discussion}

In this work we have described a means to assess the internal organizational structure of a deep neural network through the application of a task-based experimental design which is often used in studies of cognitive neuroscience. This technique showed the existence of overlapping, but distinct functional networks within the Facebook Galactica-125M model which are preferentially responsive to prompts covering political science, medical imaging, paleontology, archeology, and pathology. These networks include elements which were shown to be active across multiple repeated experiments with similar but not identical tasks. Further, by considering the intersection of elements active in a task with the elements active in a pre-computed functional network, normalized to be the percentage of active elements in that network, one can identify the performed task.

This work is a step into a field of functional assessment of deep neural networks. As suggested with the presented results, the activity of a functional subnetwork can predict the presented task of the input. The experimental "task" design could, conversely, be defined based upon a labeling of the output tokens of the LLM rather than the input prompts as described herein. The identified networks may be different from those identified in this work. Such networks may be of a different level of interest in cases where the goal is to better understand how to modulate the output of a neural network.

In such cases of assessing model performance based upon network activation, as described by the researchers at Anthropic \cite{bricken2023monosemanticity}, the activation could be used as a surrogate to assess model alignment. If a functional network associated with poor alignment is identified and that network is found to be active in a given inference task, inference could be censored or inference restarted with a different random state to achieve preferable model outputs. Thus, monitoring network activity could provide a means to predict and prevent divergent model performance.

The idea of a task-based assessment of model performance offers a unique opportunity for the retrospective analysis of model performance. For instance, if a model presents a persistent failure mode, a block designed stimulus could be created with a series of prompts that do and do not yield the failure. The network found to be active in the case of the failure mode could be of great interest in understanding and addressing the failure. In this case, the connections identified to be associated with the failure could be specifically unfrozen or otherwise targeted in fine tuning to mitigate the identified problem.

With the identification of functional networks within the large model, there is a potential to select the networks relevant to a task as a compressed model for task-specific inference. Just as cortical lesions in biological models yield function-specific deficits dependent upon their locations while preserving other functions \cite{VAIDYA2019653}, with task-specific functional networks identified, targeted lesions induced by model pruning outside of the network or networks of interest may provide a performance-sparing model compression solution.

A limitation of the presented work is its application to a model which is two to three orders of magnitude smaller than state of the art models. This limitation was intentional as a means to facilitate processing on a single modest server. There is no theoretical limitation to prevent the application of this technique to a larger model. In such a case, because each layer output value is considered independently, the assessment of an experiment could be easily be parallelized in a blockwise manor across all layer output values. Once a set of functional networks are identified, a further reduction in computational requirements could be achieved by limiting analysis to the elements included in those networks.

As noted in the introduction, this is one of many emerging techniques to understand the functioning of functional networks within deep neural networks \cite{belinkov2019analysis, olah2020zoom, elhage2022toy, bricken2023monosemanticity}. In general, other techniques have used ancillary deep learning-based models to provide this insight. Here we present a simpler method of arriving at a similar destination by utilizing general linear models in the motif of functional neuroimaging research. Arriving at similar conclusions, these techniques are clearly complimentary. While ancillary deep learning-based models can yield identification of functional sub-networks within large models, such methods effectively replace one opaque box with another. Here, if the linear models we have developed appropriately model the activity of modules within a deep neural network, a level of enhanced human interpretability is introduced. This potential benefit is at the cost of enhanced computational and storage requirements.

In the broader field, the concept of "feature visualization" could be extended to consider identified networks instead of individual neurons. While standard feature visualization techniques seek to maximize the output of a priori identified neuron outputs, optimization could be pursued to maximize correlation of layer outputs with the average regression coefficients for an identified network. Thus, existing means of understanding the workings of deep neural networks can be complimentary to the emerging understanding of deep neural networks being organized into sets of overlapping functional networks.

\section{Conclusion}

There is an emerging understanding in the consideration of deep neural networks that they are organized into overlapping functional sub-networks. These functional networks can relate to different semantic representations, offering a significant increase of potential information encoding through the superposition of activations across numerous network modules. This work shows that the techniques of functional neuroimaging can be adapted to probe these networks through a demonstration in a relatively small large language model. With this conceptual connection, the vast realm of experimental and analysis techniques of functional neuroimaging can be applied to enhance our understanding of deep neural networks. Further, with task- or outcome-specific functional networks identified within deep neural networks, this work offers the promise to identify opportunities to better align or fine tune models where these networks are mapped.


\bibliographystyle{unsrt}
\bibliography{references}

\begin{thebibliography}{10}

\bibitem{vaswani2017attention}
Ashish Vaswani, Noam Shazeer, Niki Parmar, Jakob Uszkoreit, Llion Jones, Aidan~N Gomez, {\L}ukasz Kaiser, and Illia Polosukhin.
\newblock Attention is all you need.
\newblock {\em Advances in neural information processing systems}, 30, 2017.

\bibitem{alzubaidi2021review}
Laith Alzubaidi, Jinglan Zhang, Amjad~J. Humaidi, Ayad~Qasim Al-Dujaili, and Ye~Duan.
\newblock Review of deep learning: concepts, cnn architectures, challenges, applications, future directions.
\newblock {\em Journal of Big Data}, 8(1):53, 2021.

\bibitem{brown2020language}
Tom~B Brown, Benjamin Mann, Nick Ryder, Melanie Subbiah, Jared Kaplan, Prafulla Dhariwal, Arvind Neelakantan, Pranav Shyam, Girish Sastry, Amanda Askell, Sandhini Agarwal, Ariel Herbert-Voss, Gretchen Krueger, Tom Henighan, Rewon Child, Aditya Ramesh, Daniel~M Ziegler, Jeffrey Wu, Clemens Winter, Christopher Hesse, Mark Chen, Eric Sigler, Mateusz Litwin, Scott Gray, Benjamin Chess, Jack Clark, Christopher Berner, Sam McCandlish, Alec Radford, Ilya Sutskever, and Dario Amodei.
\newblock Language models are few-shot learners.
\newblock In H.~Larochelle, M.~Ranzato, R.~Hadsell, M.F. Balcan, and H.~Lin, editors, {\em Advances in Neural Information Processing Systems}, volume~33, pages 1877--1901. Curran Associates, Inc., 2020.

\bibitem{belinkov2019analysis}
Yonatan Belinkov and James Glass.
\newblock Analysis methods in neural language processing: A survey.
\newblock {\em Transactions of the Association for Computational Linguistics}, 7:49--72, 2019.

\bibitem{wang2023aligning}
Yufei Wang, Wanjun Zhong, Liangyou Li, Fei Mi, Xingshan Zeng, Wenyong Huang, Lifeng Shang, Xin Jiang, and Qun Liu.
\newblock Aligning large language models with human: A survey, 2023.

\bibitem{zhao2023survey}
Wayne~Xin Zhao, Kun Zhou, Junyi Li, Tianyi Tang, Xiaolei Wang, Yupeng Hou, Yingqian Min, Beichen Zhang, Junjie Zhang, Zican Dong, Yifan Du, Chen Yang, Yushuo Chen, Zhipeng Chen, Jinhao Jiang, Ruiyang Ren, Yifan Li, Xinyu Tang, Zikang Liu, Peiyu Liu, Jian-Yun Nie, and Ji-Rong Wen.
\newblock A survey of large language models, 2023.

\bibitem{xi2023rise}
Zhiheng Xi, Wenxiang Chen, Xin Guo, Wei He, Yiwen Ding, Boyang Hong, Ming Zhang, Junzhe Wang, Senjie Jin, Enyu Zhou, Rui Zheng, Xiaoran Fan, Xiao Wang, Limao Xiong, Yuhao Zhou, Weiran Wang, Changhao Jiang, Yicheng Zou, Xiangyang Liu, Zhangyue Yin, Shihan Dou, Rongxiang Weng, Wensen Cheng, Qi~Zhang, Wenjuan Qin, Yongyan Zheng, Xipeng Qiu, Xuanjing Huang, and Tao Gui.
\newblock The rise and potential of large language model based agents: A survey, 2023.

\bibitem{olah2017feature}
Chris Olah, Alexander Mordvintsev, and Ludwig Schubert.
\newblock Feature visualization.
\newblock {\em Distill}, Nov 2017.

\bibitem{bricken2023monosemanticity}
Trenton Bricken, Adly Templeton, Joshua Batson, Brian Chen, Adam Jermyn, Tom Conerly, Nick Turner, Cem Anil, Carson Denison, Amanda Askell, Robert Lasenby, Yifan Wu, Shauna Kravec, Nicholas Schiefer, Tim Maxwell, Nicholas Joseph, Zac Hatfield-Dodds, Alex Tamkin, Karina Nguyen, Brayden McLean, Josiah~E Burke, Tristan Hume, Shan Carter, Tom Henighan, and Christopher Olah.
\newblock Towards monosemanticity: Decomposing language models with dictionary learning.
\newblock {\em Transformer Circuits Thread}, 2023.
\newblock https://transformer-circuits.pub/2023/monosemantic-features/index.html.

\bibitem{olah2020zoom}
Chris Olah, Nick Cammarata, Ludwig Schubert, Gabriel Goh, Michael Petrov, and Shan Carter.
\newblock Zoom in: An introduction to circuits.
\newblock {\em Distill}, March 2020.

\bibitem{elhage2022toy}
Nelson Elhage, Tristan Hume, Catherine Olsson, Nicholas Schiefer, Tom Henighan, Shauna Kravec, Zac Hatfield-Dodds, Robert Lasenby, Dawn Drain, Carol Chen, Roger Grosse, Sam McCandlish, Jared Kaplan, Dario Amodei, Martin Wattenberg, and Christopher Olah.
\newblock Toy models of superposition, 2022.

\bibitem{POLDRACK2008223}
Russell~A Poldrack.
\newblock The role of fmri in cognitive neuroscience: where do we stand?
\newblock {\em Current Opinion in Neurobiology}, 18(2):223--227, 2008.
\newblock Cognitive neuroscience.

\bibitem{doi:10.1126/science.1948051}
J.~W. Belliveau, D.~N. Kennedy, R.~C. McKinstry, B.~R. Buchbinder, R.~M. Weisskoff, M.~S. Cohen, J.~M. Vevea, T.~J. Brady, and B.~R. Rosen.
\newblock Functional mapping of the human visual cortex by magnetic resonance imaging.
\newblock {\em Science}, 254(5032):716--719, 1991.

\bibitem{doi:10.1073/pnas.89.12.5675}
K~K Kwong, J~W Belliveau, D~A Chesler, I~E Goldberg, R~M Weisskoff, B~P Poncelet, D~N Kennedy, B~E Hoppel, M~S Cohen, and R~Turner.
\newblock Dynamic magnetic resonance imaging of human brain activity during primary sensory stimulation.
\newblock {\em Proceedings of the National Academy of Sciences}, 89(12):5675--5679, 1992.

\bibitem{bandettini1992time}
PA~Bandettini, EC~Wong, RS~Hinks, RS~Tikofsky, and JS~Hyde.
\newblock Time course epi of human brain function during task activation.
\newblock {\em Magnetic Resonance in Medicine}, 25(2):390--397, Jun 1992.

\bibitem{d'esposito2003alterations}
M.~D'Esposito, L.~Deouell, and A.~Gazzaley.
\newblock Alterations in the bold fmri signal with ageing and disease: a challenge for neuroimaging.
\newblock {\em Nature Reviews Neuroscience}, 4:863--872, 2003.

\bibitem{rakhimberdina2021natural}
Zarina Rakhimberdina, Quentin Jodelet, Xin Liu, and Tsuyoshi Murata.
\newblock Natural image reconstruction from fmri using deep learning: A survey.
\newblock {\em Frontiers in Neuroscience}, 15, December 2021.

\bibitem{taylor2022galactica}
Ross Taylor, Marcin Kardas, Guillem Cucurull, Thomas Scialom, Anthony Hartshorn, Elvis Saravia, Andrew Poulton, Viktor Kerkez, and Robert Stojnic.
\newblock Galactica: A large language model for science, 2022.

\bibitem{doi:10.1073/pnas.87.24.9868}
S~Ogawa, T~M Lee, A~R Kay, and D~W Tank.
\newblock Brain magnetic resonance imaging with contrast dependent on blood oxygenation.
\newblock {\em Proceedings of the National Academy of Sciences}, 87(24):9868--9872, 1990.

\bibitem{dukart2014structure}
Juergen Dukart and Alessandro Bertolino.
\newblock When structure affects function – the need for partial volume effect correction in functional and resting state magnetic resonance imaging studies.
\newblock {\em PLOS ONE}, 9(12):e114227, December 2014.

\bibitem{LINDQUIST2009S187}
Martin~A. Lindquist, Ji~{Meng Loh}, Lauren~Y. Atlas, and Tor~D. Wager.
\newblock Modeling the hemodynamic response function in fmri: Efficiency, bias and mis-modeling.
\newblock {\em NeuroImage}, 45(1, Supplement 1):S187--S198, 2009.
\newblock Mathematics in Brain Imaging.

\bibitem{GALACTICA}
Ross Taylor, Marcin Kardas, Guillem Cucurull, Thomas Scialom, Anthony Hartshorn, Elvis Saravia, Andrew Poulton, Viktor Kerkez, and Robert Stojnic.
\newblock Galactica: A large language model for science.
\newblock 2022.

\bibitem{wolf2020transformers}
Thomas Wolf, Lysandre Debut, Victor Sanh, Julien Chaumond, Clement Delangue, Anthony Moi, Perric Cistac, Clara Ma, Yacine Jernite, Julien Plu, Canwen Xu, Teven Le~Scao, Sylvain Gugger, Mariama Drame, Quentin Lhoest, and Alexander~M. Rush.
\newblock Transformers: State-of-the-art natural language processing.
\newblock In {\em Proceedings of the 2020 Conference on Empirical Methods in Natural Language Processing: System Demonstrations}, pages 38--45, Online, 10 2020. Association for Computational Linguistics.

\bibitem{openai_chat2023}
OpenAI GPT-4.
\newblock Personal communication.
\newblock OpenAI ChatGPT, 2023.
\newblock https://platform.openai.com/.

\bibitem{nltk}
NLTK Team.
\newblock Natural language toolkit (nltk), 2023.

\bibitem{nltkbook}
Steven Bird, Ewan Klein, and Edward Loper.
\newblock {\em Natural Language Processing with Python: Analyzing Text with the Natural Language Toolkit}.
\newblock O'Reilly Media, Inc., 2009.

\bibitem{seabold2010statsmodels}
Skipper Seabold and Josef Perktold.
\newblock Statsmodels: Econometric and statistical modeling with python.
\newblock In {\em 9th Python in Science Conference}, 2010.

\bibitem{68aee965-a8a0-3e72-9f89-8d89ae91a62b}
J.~A. Nelder and R.~W.~M. Wedderburn.
\newblock Generalized linear models.
\newblock {\em Journal of the Royal Statistical Society. Series A (General)}, 135(3):370--384, 1972.

\bibitem{dunn1961multiple}
Olive~Jean Dunn.
\newblock Multiple comparisons among means.
\newblock {\em Journal of the American statistical association}, 56(293):52--64, 1961.

\bibitem{VAIDYA2019653}
Avinash~R. Vaidya, Maia~S. Pujara, Michael Petrides, Elisabeth~A. Murray, and Lesley~K. Fellows.
\newblock Lesion studies in contemporary neuroscience.
\newblock {\em Trends in Cognitive Sciences}, 23(8):653--671, 2019.

\end{thebibliography}

\end{document}